\documentclass[runningheads]{llncs}

\usepackage{eccv}
\usepackage{eccvabbrv}
\usepackage{graphicx}
\usepackage{booktabs}

\usepackage{float}
\usepackage{algorithm}
\usepackage{algorithmic}
\usepackage{multirow}
\usepackage{adjustbox}

\usepackage[accsupp]{axessibility}  % Improves PDF readability for those with disabilities.

\usepackage{hyperref}

\usepackage{orcidlink}

\begin{document}

% \title{NC-SDEdit: Noise Calibration in Diffusion Models for Content-Preserving video Refinement} 
\title{Noise Calibration: Plug-and-play Content-Preserving Video Enhancement using Pre-trained Video Diffusion Models}

\titlerunning{Noise Calibration}

\author{Qinyu Yang\inst{1} \and
Haoxin Chen\inst{2} \and
Yong Zhang\inst{2,*} \and
Menghan Xia\inst{2} \and
Xiaodong Cun\inst{2} \and
Zhixun Su\inst{1,*} \and
Ying Shan\inst{2}   }

\authorrunning{Q.~Yang et al.}

\institute{Dalian University of Technology \and
Tencent AI Lab\\
\url{https://github.com/yangqy1110/NC-SDEdit/}}
\maketitle

\begin{abstract}
% Although significant progress has been made in video generation, the synthesized videos of many methods can be enhanced for better quality. 
In order to improve the quality of synthesized videos, currently, one predominant method involves retraining an expert diffusion model and then implementing a noising-denoising process for refinement. 
Despite the significant training costs, maintaining consistency of content between the original and enhanced videos remains a major challenge.
To tackle this challenge, we propose a novel formulation that considers both visual quality and consistency of content. 
Consistency of content is ensured by a proposed loss function that maintains the structure of the input, while visual quality is improved by utilizing the denoising process of pretrained diffusion models. 
To address the formulated optimization problem, we have developed a plug-and-play noise optimization strategy, referred to as \textbf{Noise Calibration}.
By refining the initial random noise through a few iterations, the content of original video can be largely preserved, and the enhancement effect demonstrates a notable improvement. Extensive experiments have demonstrated the effectiveness of the proposed method.\footnotetext{Corresponding author}

%and its wide applicability to other SDEdit-based methods, including MS-Vid2Vid-XL and SDXL-1.0-refiner without additional training.

\end{abstract}

% \begin{figure}[tb]
%   \centering
%   \includegraphics[height=9.5cm]{fig1.pdf}
%   \caption{\textbf{Examples of Baseline (SDEdit) Demonstrations.} Using the same pre-trained T2V diffusion model, starting at the same initial denoising step(600), and using the same sampling method. However, for the video on the left, the level of added noise is too low, resulting in many brown noise points. For the video on the right, the level of added noise is too high, causing the "rubber duck", which should be a static object, to turn into a dynamic little yellow duck with its eyes and tail moving, and the sea water showing through its body in the picture.
%   }
%   \label{fig:fig1}
% \end{figure}

\begin{figure}[tb]
\centering
\begin{minipage}[t]{0.48\textwidth}
% \vspace{-1cm} 
\centering
\includegraphics[width=5.2cm]{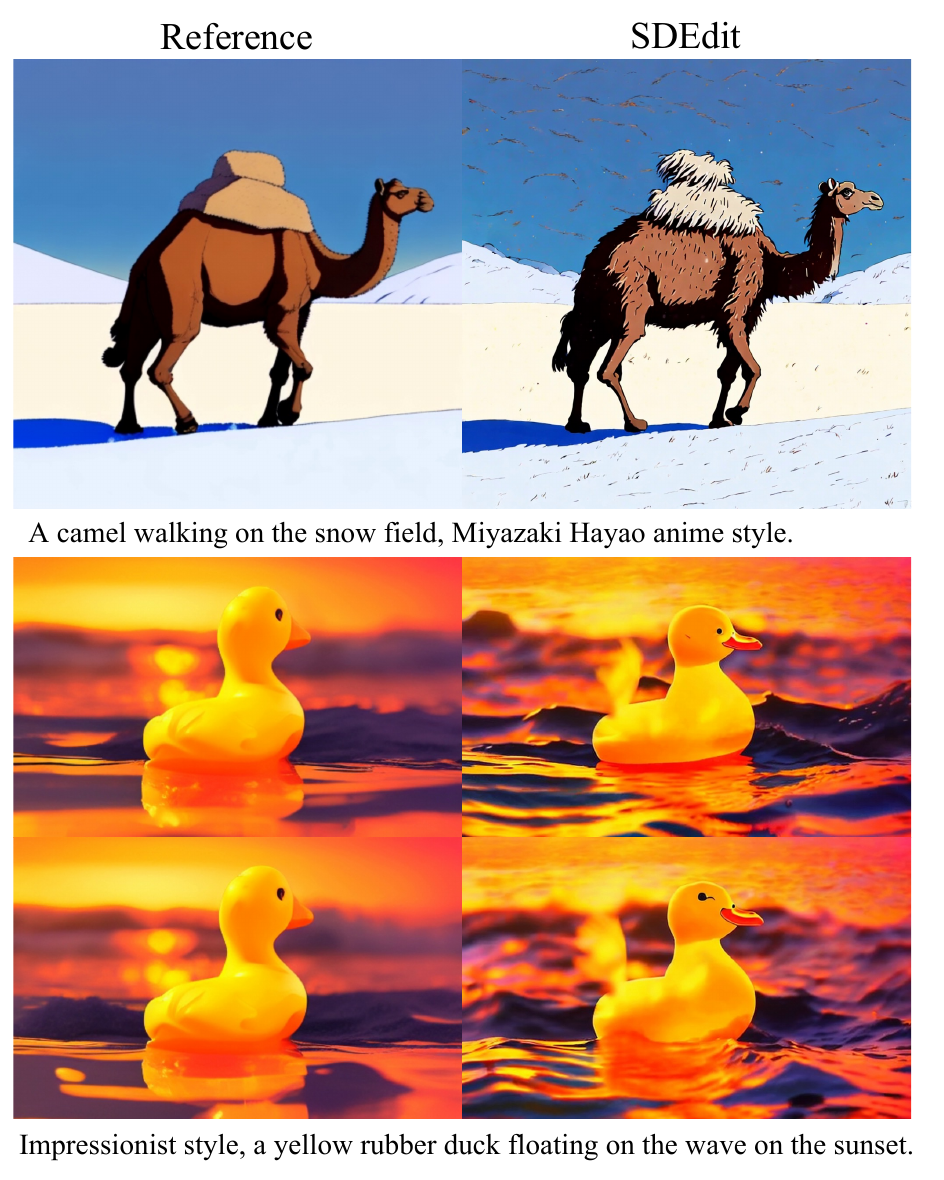}
\caption{Examples demonstrating video enhancement based on SDEdit}
\label{fig:fig1}
\end{minipage}
\begin{minipage}[t]{0.48\textwidth}
\centering
\includegraphics[width=6cm]{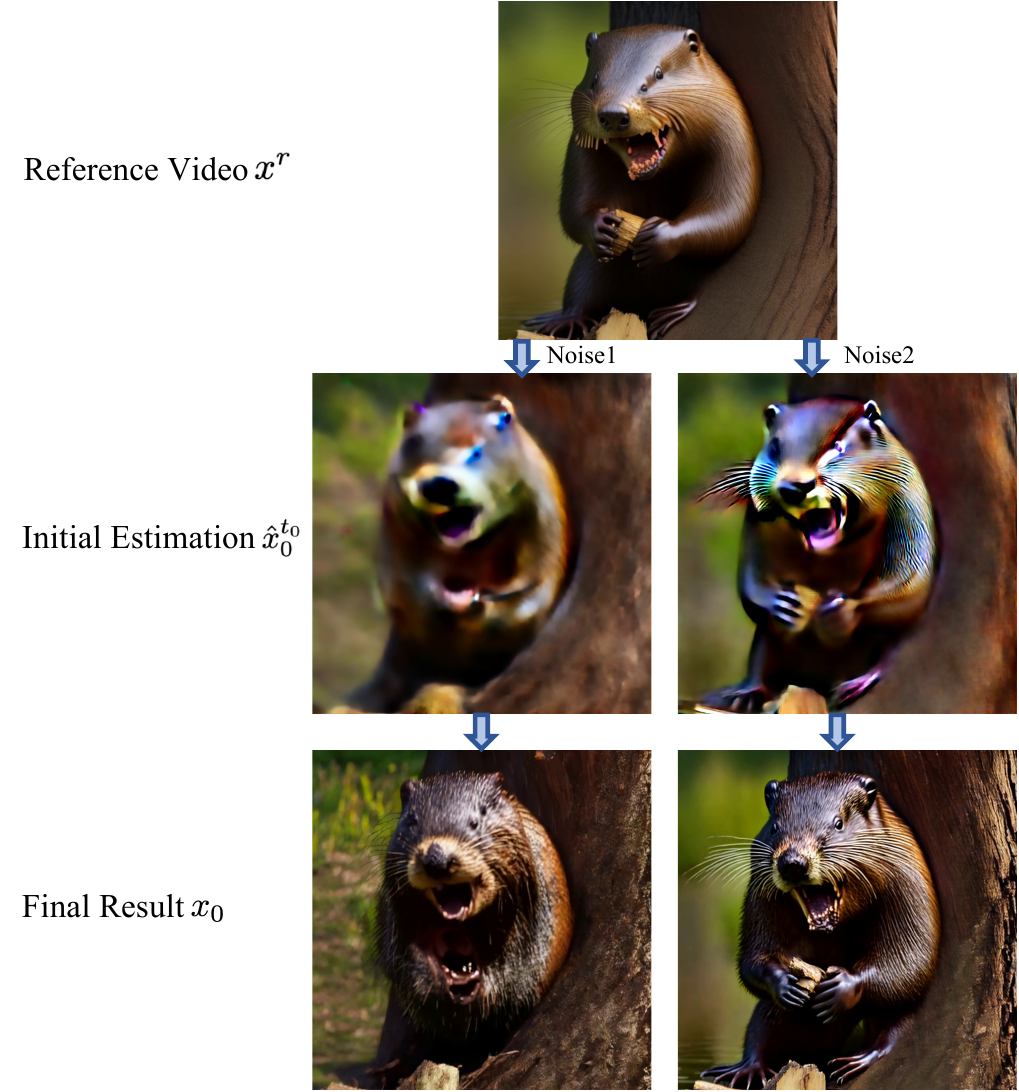}
\caption{Decomposition of the video enhancement process based on SDEdit
}
\label{fig:fig2}
\end{minipage}
\end{figure}

\section{Introduction}
\label{sec:intro}

Recently, diffusion models have emerged as a distinct type of generative models, in contrast to traditional Generative Adversarial Networks (GANs) \cite{goodfellow2020generative} and Variational Autoencoders (VAEs) \cite{kingma2013auto}. These models \cite{ho2020denoising,rombach2022high} have demonstrated superior performance across a wide range of applications. In particular, visual synthesis has significantly benefited from the development of diffusion models. A popular subset of methods \cite{chen2023videocrafter1,he2022latent,wang2023modelscope,zhou2022magicvideo,miech2019howto100m,rombach2022high,wang2023lavie,wang2023videofactory,yu2023video,mei2023vidm,an2023latent} leverages pre-trained Text-to-Image (T2I) models and incorporates additional temporal blocks to extend video generation capabilities. However, obtaining results with both accurate semantics and high visual quality through a single inference often proves challenging. Typically, a low-quality video with diverse and semantically accurate motions is generated from the base model, and then an expert model is retrained to implement a noising-denoising process, as pioneered by SDEdit\cite{meng2021sdedit}, to refine the generated video. The expert model is further trained using high-quality videos with earlier noise scales, aiming to amplify the model's attention to spatio-temporal details. This approach is anticipated to further improve the spatio-temporal continuity and clarity of the video, effectively addressing artifacts in both time and space dimensions.

However, despite the substantial resources invested in retraining, the well-established structures in the original video are often disrupted during the process of quality enhancement. Consequently, we believe that solely utilizing pre-trained Text-to-Video (T2V) diffusion models for content-preserving video enhancement is a worthwhile research direction. This T2V model does not need to generate videos with accurate semantic motions; however, it is required to produce videos with high visual quality. By only using SDEdit for video enhancement based on pre-trained T2V models, it is common for the spatial structure of the original video to be disrupted when the initial denoising step is set sufficiently large to achieve satisfactory quality enhancement, as shown in \cref{fig:fig1}. For instance, in the "camel" example, while the quality is enhanced, brown camel hair appears to be flying across the sky. In the "rubber duck" example, the duck transforms from a static object into a dynamic creature, and the sunlight is transformed into a swaying tail.

To address this challenge, we propose a novel formulation for video enhancement that ensures  both quality improvement and consistency with the content of original video. Specifically, noise is first introduced to disrupt a given video, which is then gradually eliminated based on a pre-trained model. By leveraging the ability of the pre-trained model to generate high-quality videos, we enhance the quality of the original video. To ensure consistency, we propose an additional loss function that constrains the content loss between the enhancement result and the original video. To solve this optimization problem, we provide a simple but effective solution, \textbf{Noise Calibration}. By refining the initial random noise only 1-3 times before adding it to the original video, we can largely preserve the content of original video and significantly improve the enhancement effect.

A significant amount of theoretical analysis and experiments demonstrate that our method can effectively preserve the content of videos before and after enhancement when using pre-trained T2V models for video enhancement. Furthermore, this approach can conveniently serve as a plug-in to enhance the performance of state-of-the-art visual refinement models. We summarize our contributions as follows:
  \begin{itemize}
  \item We introduce a novel formulation for video enhancement based on diffusion models, which focuses not only on improving quality but also on maintaining consistency of content with the original video.
  \item We propose a concise yet effective content-preserving strategy for video enhancement, called \textbf{Noise Calibration}, which only requires calibrating the initial random noise, without any additional fine-tuning or operations.
  \item Extensive quantitative and qualitative experiments demonstrate that Noise Calibration can be effectively applied to video enhancement and various tasks based on SDEdit, achieving more controllable image/video generation.
\end{itemize}

\section{Related Work}

The aim of this study is to investigate how to maintain consistency of content between the enhancement results and the original video while performing video enhancement based on SDEdit. Consequently, we will briefly review relevant domains in this section to facilitate a better understanding.

\subsection{Video Diffusion Models}
The recent emergence of diffusion models \cite{sohl2015deep,ho2020denoising,song2019generative,song2020score,dhariwal2021diffusion,nichol2021improved,song2020denoising,watson2022learning} as a type of generative model \cite{kingma2019introduction,oussidi2018deep,lecun2006tutorial,ngiam2011learning,creswell2018generative,goodfellow2020generative,rezende2015variational} has significantly advanced the field of T2I generation \cite{sohl2015deep,ho2020denoising,song2019generative,song2020score,dhariwal2021diffusion,nichol2021improved,song2020denoising,watson2022learning,saharia2022photorealistic,song2020improved,nichol2021glide,song2021maximum,sinha2021d2c,vahdat2021score,pandey2021vaes,bao2022analytic,dockhorn2021score,liu2022compositional,jiang2022text2human}. These models have also demonstrated potential in various tasks, such as image-to-image translation \cite{choi2021ilvr,meng2021sdedit,saharia2022palette,zhao2022egsde,wang2022pretraining,li2022vqbb,wolleb2022swiss}, image super-resolution \cite{yue2023resshift,saharia2022image,xia2023diffir,wang2023exploiting,chen2023iterative,luo2023image}, image inpainting \cite{yang2023paint,nichol2021glide,avrahami2022blended,avrahami2023blended,lu2023tf}, and image editing \cite{ma2024followyourpose,hertz2022prompt,parmar2023zero,mokady2023null,brooks2023instructpix2pix,ma2024followyourclick,tumanyan2023plug,kawar2023imagic,ma2024followyouremoji}, among others.

Text-to-video synthesis is a complex and challenging task with significant practical implications, as it aims to generate relevant videos from textual descriptions. Early approaches \cite{balaji2019conditional,skorokhodov2022stylegan,tulyakov2018mocogan,wang2020g3an,wang2020imaginator,wang2023styleinv} primarily utilized GANs, which unfortunately resulted in subpar video quality. As a pioneering work introducing diffusion models to the field of video generation, Video Diffusion Models \cite{2022Video} adopted the 3D U-Net from \cite{cciccek20163d}, achieving impressive results in both unconditional and text-conditional video generation tasks. Subsequently, to reduce training costs, a significant number of studies \cite{chen2023videocrafter1,he2022latent,wang2023modelscope,zhou2022magicvideo,miech2019howto100m,rombach2022high,wang2023lavie,wang2023videofactory,yu2023video,mei2023vidm,an2023latent} have extended pre-trained image diffusion models to video and learned Video Diffusion Models in latent or hybrid-pixel-latent spaces.

Given the complexity of video generation, limited computational resources, and the scarcity of high-quality video data, cascade models \cite{ho2022imagen,hu2022make,zhang2023show,zhang2023i2vgen,wang2023lavie} have emerged as the mainstream paradigm, adopting a divide-and-conquer approach to tackle these challenges. Specifically, a cascade model typically comprises three components: a base model, a frame interpolation model, and a refinement model. Based on the base model trained with a large number of low-resolution videos, we can generate well-structured low-resolution videos. Subsequently, the frame interpolation model enhances the video's continuity by adding frames. Finally, a refinement model is employed to further improve the spatio-temporal continuity and clarity of the video.

\subsection{Refinement Models of Video Diffusion Models}
\label{sec:22}
As mentioned above, refinement models play a crucial role in determining the final quality of generated videos. In this section, we will review the methods employed by existing refinement models. Refinement models differ from conventional super-resolution models, which solely focus on increasing the resolution. A critical aspect of refinement models is their ability to refine the videos generated by the base model, which might lack sufficient details, by adding appropriate details to enhance the overall quality. There are few existing methods, which can primarily be categorized into two approaches: methods based on SR3 \cite{saharia2022image}, which resemble traditional super-resolution techniques, and methods based on SDEdit \cite{meng2021sdedit}, which lean towards further generation.

As the first diffusion-based Super Resolution method, SR3 \cite{saharia2022image} incorporates the low-resolution (LR) image as an additional input to the denoising network, constructing a conditional denoising network. To further enhance visual quality of generated videos and increase spatial resolution, Lavie \cite{wang2023lavie} utilizes SD-x4-Upscaler \cite{rombach2022high} as a prior and incorporates an additional temporal dimension, enabling temporal processing within the diffusion UNet. Although the additional initial video embedding greatly ensures consistency of content before and after enhancement, it also restricts the refinement capabilities of diffusion models.

SDEdit \cite{meng2021sdedit} is a pioneering approach that achieves editing through iterative denoising via a stochastic differential equation (SDE). Initially designed to address the Stroke Painting to Image problem, SDEdit's impressive scalability has since facilitated advancements in various other fields \cite{wu2023latent,ye2023affordance,podell2023sdxl}. In the realm of video generation, both Show-1 \cite{zhang2023show} and Modelscope \cite{zhang2023i2vgen} have successfully implemented SDEdit to enhance the quality of videos produced by their base models. To amplify the refinement model's focus on spatio-temporal details, they specifically train it on low noise scales, using high-resolution videos. Although this method possesses a stronger generative capability and can enrich video details more effectively, the randomness inherent in the generation process may lead to the final video deviating from the original content or even damaging well-established structures. To address this issue, we propose a training-free and plug-and-play method that aims to enhance the consistency of content between the final video and the original video without compromising the refinement quality. This method is versatile, compatible with both pre-trained T2V models and expert models retrained on a low noise scale.

% \subsection{Subsequent Work of SDEdit}
Despite SDEdit's wide application across various tasks, the trade-off between realism and fidelity often falls short of user expectations in practical applications, as noted in several studies\cite{yang2023eliminating,kim20233d,hachnochi2023cross,zhang2023forgedit,ahn2023interactive,brack2023ledits++,jimenez2023mixture,mishra2023syncdr}. Despite this, there is a noticeable lack of research in this area. Peng et al. \cite{peng2023diffusion} proposed a method that uses source semantics to guide the generation process, aiming to enhance the consistency between the source and target domain content in SDEdit's image translation tasks. However, this method's effectiveness is contingent on the accuracy of semantic maps, and its applicability in the latent space is yet to be assessed. Singh et al. \cite{singh2023high} also concentrated on optimizing the subsequent sampling process with the goal of enhancing the realism of the editing results for stroke painting to image. However, a universally applicable method has not been established yet. Our approach, while primarily designed for content-preserving video enhancement, can be easily adapted for other tasks based on SDEdit. Additionally, it is worth noting that our work is partially inspired by ILVR \cite{choi2021ilvr}, which refines each generation step with low-frequency component of purturbed reference image for controlling the generation of unconditional Diffusion Models.

\section{Proposed Methods}
% Given a low-quality video with diverse and semantically accurate motions, our objective is to enrich its content by adding details and textures, ultimately improving the overall video quality. Technically speaking, this challenge can be described as a specific type of video-to-video translation, which places a high demand on visual consistency. This challenge can be technically characterized as a specific form of video-to-video translation, which requires a strong emphasis on visual consistency. As a pioneering approach, SDEdit is capable of delivering satisfactory enhancement results; therefore, our primary focus lies in maintaining consistency of content throughout the process.

\subsection{Preliminaries}
Diffusion models start from the given image $x_0$, and then progressively add Gaussian Noise $\epsilon_t \sim \mathcal{N}(0,1)$. This transformation yields $x_t$ in each timestep $t$, which can be directly computed as:
\begin{equation}
x_t=\sqrt{\bar{\alpha}_t} x_0+\sqrt{1-\bar{\alpha}_t} \epsilon_t,
\end{equation}
where $\bar{\alpha}_t$ 
  represents the diffusion schedule parameters following a given sequence $0 = \bar{\alpha}_T < \bar{\alpha}_{T-1}... < \bar{\alpha}_1 < \bar{\alpha}_0 = 1$.
During inference, diffusion models can synthesize new image by starting with a random noise sample $x_T \sim \mathcal{N}(0,1)$ and iteratively denoising it. Given a noised image $x_t$ at timestep $t$, the model predicts the next-step $x_{t-1}$ as follows:
\begin{equation}
\begin{aligned}
& p_\theta\left(x_{t-1} \mid x_t\right)=\mathcal{N}\left(\mu_\theta\left(x_t, t\right), \sigma_t \mathbf{I}\right), \\
& \mu_\theta\left(x_t, t\right)=\frac{1}{\sqrt{\alpha_t}}\left(x_t-\frac{1-\alpha_t}{\sqrt{1-\bar{\alpha}_t}} \epsilon_\theta\left(x_t, t\right)\right),
\end{aligned}
\end{equation}
where $\epsilon_\theta$ represents a neural network trained to predict the noise at each step. At each timestep, the noiseless image 
$\hat{x}_0^t$ can be approximated as:
\begin{equation}
\hat{x}_0^t=\frac{x_t}{\sqrt{{\bar{\alpha}}_t}}-\frac{\sqrt{1-\bar{\alpha}_t} \epsilon_\theta\left(x_t, t\right)}{\sqrt{\bar{\alpha}_t}} .
\label{eq:eq3}
\end{equation}

As a pioneering method, SDEdit \cite{meng2021sdedit}  introduces a reference image $x^{r}$ to initialize the denoising process at an intermediate step $t_0 \in [0, T]$. This initialization takes the form $x_{t_0} \sim \mathcal{N}\left(\sqrt{\bar{\alpha}_{t_0}} x^{r},\left(1-\bar{\alpha}_{t_0}\right) \mathbb{I}\right)$. The choice of $t_0$ represents a trade-off between realism $(t_0 \approx T)$, understood as producing images in line with the training distribution $p^{\ast}(x)$, and faithfulness $(t_0 \approx 0)$, emphasizing similarity with the reference image $x^{r}$. %The video enhancement method based on SDEdit is illustrated in \cref{alg:example}.

% \begin{algorithm}[tb]
%    \caption{SDEdit for video enhancement}
%    \label{alg:example}
% \begin{algorithmic}
%    \STATE {\bfseries Input:} reference video $x^{r}$, initial denoising step $t_0$, diffusion model $\epsilon_\theta(x_t, t)$
%    \STATE $\epsilon_{t_0} \sim \mathcal{N}(0,1)$
%    \STATE $x_{t_0}=\sqrt{\bar{\alpha}_{t_0}} x^{r}+\sqrt{1-\bar{\alpha}_{t_0}} \epsilon_{t_0}$
%    \FOR{$t=t_0$ {\bfseries to} $1$}
%         \STATE $\epsilon_{t} \sim \mathcal{N}(0,1)$
%         \STATE $\boldsymbol{x}_{t-1}=\sqrt{\bar{\alpha}_{t-1}} \left(\frac{\boldsymbol{x}_t-\sqrt{1-\bar{\alpha}_t} \epsilon_\theta(x_t, t)}{\sqrt{\bar{\alpha}_t}}\right) + \sqrt{1-\bar{\alpha}_{t-1}-\sigma_t^2} \cdot \epsilon_\theta(x_t, t)+\sigma_t \epsilon_t$
        
%    \ENDFOR
% \end{algorithmic}
% \end{algorithm}

\subsection{Formulation of Content-Preserving Video Enhancement}
As illustrated in \cref{fig:fig1}, when only using random noise $\epsilon_{t_0}$ to perturb the reference video $x^{r}$ and subsequently applying quality enhancement based on a pre-trained video model $\epsilon_\theta(x_t, t)$, achieving satisfactory enhancement often leads to unintended alterations in content. To tackle this issue, we propose a novel formulation for video enhancement that prioritizes not only visual quality but also imposes constraints on the content loss between the enhanced result $x_0$ and the reference video $x^r$, as illustrated below:
\begin{equation}
\begin{array}{ll}\min_{\epsilon_{t_0}} & dist\left(x_0, x^{r}\right) \\ \text { s.t. } & x_0 \sim P_\theta\left(x \mid x^r, \epsilon_{t_0}\right)\end{array}.
\label{eq:eq4}
\end{equation}
% where, $x_0$ represents the enhancement result of the reference video $x^{r}$ based on pre-trained video model $\epsilon_\theta(x_t, t)$, and $dist$ denotes the consistency distance metric between $x_0$ and $x^{r}$. The smaller the value, the higher the consistency of content between the two.

Using Noise1 and Noise2 from \cref{fig:fig2} as examples, given reference video $x^{r}$ and pre-trained video model $\epsilon_\theta(x_t, t)$, various initial noises $\epsilon_{t_0}$ will generate diverse enhanced videos $x_0$ based on the training distribution $P_\theta$ of the pre-trained video model $\epsilon_\theta(x_t, t)$. The optimization goal is to identify a more appropriate initial noise $\epsilon_{t_0}$ that effectively minimizes the content loss, $dist\left(x_0, x^{r}\right)$, between the reference video $x^{r}$ and the enhanced video $x_0$. A smaller value of $dist\left(x_0, x^{r}\right)$ indicates a higher consistency of content between the two videos.

To derive a specific optimizable form of $dist\left(x_0, x^{r}\right)$, we decompose the video enhancement process based on SDEdit. Specifically, the reference video $x^{r}$ is combined with random noise $\epsilon_{t_0}$ corresponding to the selected initial denoising step $t_0$, using the following formula:
\begin{equation}
x_{t_0}=\sqrt{\bar{\alpha}_{t_0}} x^{r}+\sqrt{1-\bar{\alpha}_{t_0}} \epsilon_{t_0}, \epsilon_{t_0} \sim \mathcal{N}(0,1).
\label{eq:eq5}
\end{equation}

Based on \cref{eq:eq3}, the initial estimation $\hat{x}_0^{t_0}$ of the enhancement result at step $t_0$ can be expressed as:
\begin{equation}
\hat{x}_0^{t_0}=\frac{x_{t_0}-\sqrt{1-\bar{\alpha}_{t_0}} \epsilon_\theta\left(x_{t_0}, t_0\right)}{\sqrt{{\bar{\alpha}}_{t_0}}}.
\label{eq:eq10}
\end{equation}

In accordance, $x_{t_0}$ is decomposed using the following equation:
\begin{equation}
x_{t_0} = \sqrt{\bar{\alpha}_{t_0}} \hat{x}_0^{t_0} + \sqrt{1-\bar{\alpha}_{t_0}} \epsilon_{\theta}(x_{t_0}, {t_0}).
\label{eq:eq6}
\end{equation}

Subsequently, the noise video $x_{t_0-1}$ at steps $t_0-1$ is obtained as:
\begin{equation}
x_{t_0-1} = \sqrt{\bar{\alpha}_{t_0-1}} \hat{x}_0^{t_0} + \sqrt{1-\bar{\alpha}_{t_0-1}} \epsilon_{t_0-1},
\label{eq:eq7}
\end{equation}
where, if the DDIM \cite{2021Denoising} sampling method is utilized, $\epsilon_{t_0-1}$ should be $\epsilon_{\theta}(x_{t_0}, {t_0})$. Subsequently, noise video $x_t$ undergoes progressive denoising, and the corresponding estimation $\hat{x}_0^t$ for the enhancement result is gradually refined until the final enhancement result $x_0$ is obtained.

%It can be observed that, compared to $x_{t_0}$, the proportion of $\hat{x}_0^{t_0}$ in $x_{t_0-1}$ will increase.

% As shown in \cref{fig:fig2}, the above process can be divided into two steps. First, reference video $x^{r}$ is subjected to random noise, Noise1, based on \cref{eq:eq5} and then passed through diffusion model $\epsilon_\theta(x_t, t)$ to obtain the initial estimation $\hat{x}_0^{t_0}$ for the enhancement result based on \cref{eq:eq10}. A sufficiently large noise addition leads to a significant loss of consistency of content between $x^{r}$ and $\hat{x}_0^{t_0}$. Subsequently, during the progressive denoising process, the initial estimation $\hat{x}_0^{t_0}$ of the enhancement result $x_0$ is gradually refined until the final enhancement result $x_0$ is obtained. However, there is almost no consistency loss in this process.%, because, compared to the previous step $t$, the proportion of $\hat{x}_0^{t}$ affecting the content in each step of the noise video $x_{t-1}$ will be greater, as shown in \cref{eq:eq6,eq:eq7}. 

As illustrated in \cref{fig:fig2}, the primary content loss occurs between the initial estimation $\hat{x}_0^{t_0}$ and the reference video $x^r$ during the enhancement process. Therefore, we propose to measure $dist\left(x_0, x^{r}\right)$ by assessing the difference between the low-frequency components of $\hat{x}_0^{t_0}$ and $x^{r}$, following the decomposition method outlined in \cite{si2023freeu}:
\begin{equation}
	\begin{split}
	\quad x^{r} &= f_{l}^{\nu}(x^{r}) + f_{h}^{\nu}(x^{r}), \\
	\hat{x}_0^{t_0} &= f_{l}^{\nu}(\hat{x}_0^{t_0}) + f_{h}^{\nu}(\hat{x}_0^{t_0}),
	\end{split}
 \label{eq:eq8}
\end{equation}
where $f_{l}^\nu$ denotes the low-frequency component, $f_{h}^\nu$ denotes the high-frequency component, and the threshold frequency $\nu$ lies between 0 and 1, defining the boundary between high and low frequencies.

Based on the analysis above, we redefine our formulation as follows:
\begin{equation}
\begin{array}{ll}\min_{\epsilon_{t_0}} & ||f_{l}^\nu(x^{r})-f_{l}^\nu(\hat{x}_0^{t_0})|| \\ \text { s.t. } & \hat{x}_0^{t_0} \sim \hat{P}_\theta^{t_0}\left(x \mid x^r, \epsilon_{t_0}\right)\end{array},
\label{eq:eq9}
\end{equation}
where $\hat{x}_0^{t_0}$ represents the initial estimation of the enhancement result $x_0$.

\subsection{Noise Calibration}
In order to solve the formulation defined in the previous section, we propose a simple and effective optimization method called \textbf{Noise Calibration}, which essentially obtains a more suitable initial noise through 1-3 iterations. Specifically, by combining \cref{eq:eq5,eq:eq6,eq:eq8}, we obtain:
\begin{equation}
\sqrt{\bar{\alpha}_{t_0}} (f_{l}^\nu(x^{r})-f_{l}^\nu(\hat{x}_0^{t_0})) = \sqrt{1-\bar{\alpha}_{t_0}} (\epsilon_{\theta}(x_{t_0}, {t_0})-\epsilon_{t_0}) + \sqrt{\bar{\alpha}_{t_0}} (f_{h}^\nu(\hat{x}_0^{t_0}) - f_{h}^\nu(x^{r})).
\end{equation}

This implies that,
\begin{equation}
\min_{\epsilon_{t_0}} ||f_{l}^\nu(x^{r})-f_{l}^\nu(\hat{x}_0^{t_0})|| \equiv \min_{\epsilon_{t_0}} ||\epsilon_{\theta}(x_{t_0}, {t_0})-\epsilon_{t_0}+\frac{\sqrt{\bar{\alpha}_{t_0}}}{\sqrt{1-\bar{\alpha}_{t_0}}}(f_{h}^\nu(\hat{x}_0^{t_0}) - f_{h}^\nu(x^{r}))||.
\label{eq:eq12}
\end{equation}

% Moreover, $\epsilon_{\theta}(x_{t_0}, {t_0})-\epsilon_{t_0}+\frac{\sqrt{\bar{\alpha}_{t_0}}}{\sqrt{1-\bar{\alpha}_{t_0}}}(f_{h}^\nu(\hat{x}_0^{t_0}) - f_{h}^\nu(x^{r}))=0$ implies that,
% \begin{equation}
% \begin{aligned}
% \underline{\epsilon_{t_0}} &= \epsilon_{\theta}(x_{t_0}, {t_0}) + \frac{\sqrt{\bar{\alpha}_{t_0}}}{\sqrt{1-\bar{\alpha}_{t_0}}} (f_{h}^\nu(\hat{x}_0^{t_0})-f_{h}^\nu(x^{r})) \\
% &=\underline{\epsilon_{t_0}}+ \frac{\sqrt{\bar{\alpha}_{t_0}}}{\sqrt{1-\bar{\alpha}_{t_0}}} (f_{l}^\nu(x^{r})-f_{l}^\nu(\hat{x}_0^{t_0}))
% \end{aligned}
% \label{eq:eq12}
% \end{equation}

The goal of ILVR \cite{choi2021ilvr} is to generate an image $x \in \left\{x: \phi(x)=\phi(y)\right\}$ based on a diffusion model, given a reference image $y$. Here, $\phi(\cdot)$ denotes a linear low-pass filtering operation, a sequence of downsampling and upsampling. This goal is achieved by ensuring  $\phi(x_t)=\phi(y_t)$ in the denoising process through:
\begin{equation}
x_{t} \leftarrow x_{t}+\phi\left(y_{t}\right)-\phi\left(x_{t}\right).
\label{eq:ilvr}
\end{equation}

Motivated by this insight, we employ Fixed Point Iteration based on \cref{eq:eq12} to optimize the initial noise as:
\begin{equation}
\underline{\epsilon_{t_0}} \leftarrow \epsilon_{\theta}(\sqrt{\bar{\alpha}_{t_0}} x^{r}+\sqrt{1-\bar{\alpha}_{t_0}} \underline{\epsilon_{t_0}}, {t_0})+\frac{\sqrt{\bar{\alpha}_{t_0}}}{\sqrt{1-\bar{\alpha}_{t_0}}}(f_{h}^\nu(\hat{x}_0^{t_0}) - f_{h}^\nu(x^{r})).
\end{equation}

% Similarly, to achieve our optimization goal, we construct an iterative process to calibrate the initial noise as:
% \begin{equation}
% \epsilon_{t_0} \leftarrow \epsilon_{t_0}+ \frac{\sqrt{\bar{\alpha}_{t_0}}}{\sqrt{1-\bar{\alpha}_{t_0}}} (f_{l}^\nu(x^{r})-f_{l}^\nu(\hat{x}_0^{t_0})).
% \end{equation}

\begin{algorithm}[tb]
   \caption{SDEdit with Noise Calibration for video enhancement}
   \label{alg:alg2}
\begin{algorithmic}
   \STATE {\bfseries Input:} reference video $x^{r}$, initial denoising step $t_0$, diffusion model $\epsilon_\theta(x_t, t)$, iteration steps $N$, threshold frequency $\nu$
   \STATE $\epsilon_{t_0} \sim \mathcal{N}(0,1)$
   \FOR{$n=1$ {\bfseries to} $N$}
   \STATE $x_{t_0}=\sqrt{\bar{\alpha}_{t_0}} x^{r}+\sqrt{1-\bar{\alpha}_{t_0}} \epsilon_{t_0}$
   \STATE $\hat{x}_0^{t_0}=(x_{t_0} - \sqrt{1-\bar{\alpha}_{t_0}} \epsilon_\theta\left(x_{t_0}, {t_0}\right))  / \sqrt{\bar{\alpha}_{t_0}}$
   \STATE $\epsilon_{t_0} = \epsilon_{\theta}(x_{t_0}, {t_0}) + \frac{\sqrt{\bar{\alpha}_{t_0}}}{\sqrt{1-\bar{\alpha}_{t_0}}} (f_{h}^\nu(\hat{x}_0^{t_0})-f_{h}^\nu(x^{r}))$
   \ENDFOR
   \STATE $x_{t_0}=\sqrt{\bar{\alpha}_{t_0}} x^{r}+\sqrt{1-\bar{\alpha}_{t_0}} \epsilon_{t_0}$
   \FOR{$t=t_0$ {\bfseries to} $1$}
        \STATE $\epsilon_{t} \sim \mathcal{N}(0,1)$
        \STATE $\boldsymbol{x}_{t-1}=\sqrt{\bar{\alpha}_{t-1}} \left(\frac{\boldsymbol{x}_t-\sqrt{1-\bar{\alpha}_t} \epsilon_\theta(x_t, t)}{\sqrt{\bar{\alpha}_t}}\right) + \sqrt{1-\bar{\alpha}_{t-1}-\sigma_t^2} \cdot \epsilon_\theta(x_t, t)+\sigma_t \epsilon_t$
   \ENDFOR
\end{algorithmic}
\end{algorithm}

Essentially, during each iteration, a replacement on the low-frequency domain resembling LIVR \cite{choi2021ilvr} is performed on $x_{t_0}$ as:
\begin{equation}
x_{t_0} \leftarrow x_{t_0}+\sqrt{\bar{\alpha}_{t_0}}(f_{l}^\nu(x^{r})-f_{l}^\nu(\hat{x}_0^{t_0})).
\end{equation}

% \begin{align}
% \epsilon_{t_0} &\leftarrow \epsilon_{t_0}+ \frac{\sqrt{\bar{\alpha}_{t_0}}}{\sqrt{1-\bar{\alpha}_{t_0}}} (f_{l}^\nu(x^{r})-f_{l}^\nu(\hat{x}_0^{t_0})) \label{eq:diedai1}. \\
% \hat{x}_0^{t_0} &\leftarrow \hat{x}_0^{t_0}+f_{l}^\nu(x^{r})-f_{l}^\nu(\hat{x}_0^{t_0}) \label{eq:diedai2}.
% \end{align}

After obtaining the calibrated noise through 1-3 iterations, we re-add noise to the reference video and enhance video quality using pre-trained video models, as shown in \cref{alg:alg2}, referred to as NC-SDEdit. As illustrated in \cref{fig:fig2}, Noise2, being a calibrated version of Noise1, results in a a superior initial estimation $\hat{x}_0^{t_0}$, thereby effectively achieving content-preserving video enhancement.

\begin{figure}[tb]
  \centering
  \includegraphics[height=8cm]{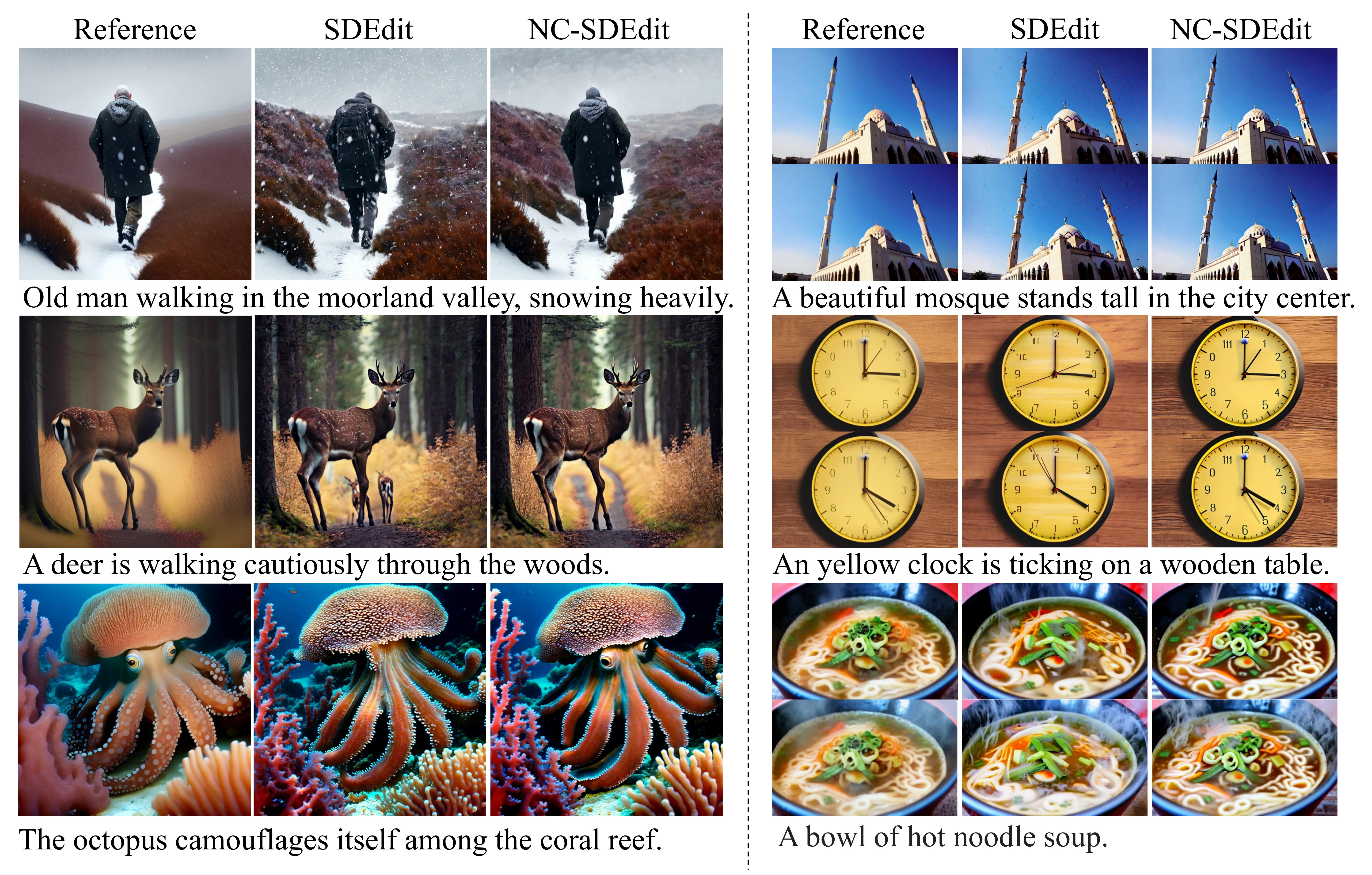}
  \caption{Visual comparisons of video enhancement based on VideoCrafter \cite{chen2023videocrafter1}}
  \label{fig:zongshow}
\end{figure}

\section{Experiments}
\subsection{Experiments Setup}
\textbf{Setting up.} We conduct our experiments using an open-source T2V diffusion model, VideoCrafter(576$\times$1024)\cite{chen2023videocrafter1}, known for its superior visual quality, albeit with limitations in semantic understanding. We use SDEdit as a benchmark, while introducing our approach, NC-SDEdit,  which essentially incorporates Noise Calibration into SDEdit. The initial denoising step, iteration steps, and threshold frequency are set to $t_0=600,N=3,\nu=1.0$ as default.

\noindent \textbf{Dataset and Metrics.}  We utilize a reference set consisting of 700 videos with a resolution of 320$\times$512, generated by Lavie\cite{wang2023lavie}, along with their corresponding texts from EvalCrafter\cite{liu2023evalcrafter} for quantitative evaluation. Firstly, to evaluate the consistency of content, we report the $\mathrm{MSE_l}$ (MSE on the low-frequency domain with $\nu$=0.5), MSE and SSIM \cite{wang2004image} between the enhancement results and the reference videos. Secondly, to assess the visual quality of the enhanced results, we also report the
state-of-the-art video quality assessment metric, DOVER \cite{wu2023exploring}. The image quality metric CLIP-IQA \cite{wang2023exploring} is also used to assist in the evaluation. Finally, spatial frequency ($\mathrm{SF}$) can measure the gradient distribution thus revealing the detail and texture of the video frame. Therefore, we use $\mathrm{D_{SF}} = \mathrm{SF}(x) - \mathrm{SF}(x^r)$ to measure whether video details have been enhanced.

\begin{table}[tb]
  \caption{Quantitative comparisons based on VideoCrafter \cite{chen2023videocrafter1}
  }
  \label{tab:headings}
  \centering
\begin{tabular}{l|ccc|cc|c}
\hline Method& $\mathrm{MSE_l}$$\downarrow$ & MSE$\downarrow$ & SSIM$\uparrow$ & DOVER$\uparrow$ & CLIP-IQA$\uparrow$& $\mathrm{D_{SF}}$$\uparrow$ \\
\hline SDEdit& 4.3447 & 0.7600 & 0.6464 & 60.17 & $\mathbf{0.4482}$& 0.0527 \\
Ours($N$=1) & $\underline{2.9201}$ & $\underline{0.6546}$ & $\underline{0.6998}$ & $\underline{60.62}$ & $\underline{0.4471}$& 0.0531 \\
Ours($N$=2)& $\mathbf{2.8039}$ & $\mathbf{0.6506}$ & $\mathbf{0.7040}$ & $\mathbf{62.71}$ & 0.4400& 0.0554 \\
Ours($N$=3) & 2.9540 & 0.6638 & 0.6971 & 62.45 & 0.4387& $\underline{0.0584}$ \\
Ours($N$=10) & 4.6107 & 0.7570 & 0.6209 & 54.42 & 0.3873 & $\mathbf{0.0741}$\\
\hline
\end{tabular}
\end{table}

\begin{table}[tb]
  \caption{User study of human preference
  }
  \label{tab:userstudy}
  \centering
\begin{tabular}{lcccc}
\hline Method & Consistency & Visual Quality & Texture \\
\hline SDEdit & 13.89\% & 26.74\% & 24.65\% \\
NC-SDEdit & \textbf{86.11\%} & \textbf{73.26\%} & \textbf{75.35}\% \\
\hline
\end{tabular}
\end{table}

\begin{figure}[tb]
  \centering
  \includegraphics[height=2.3cm]{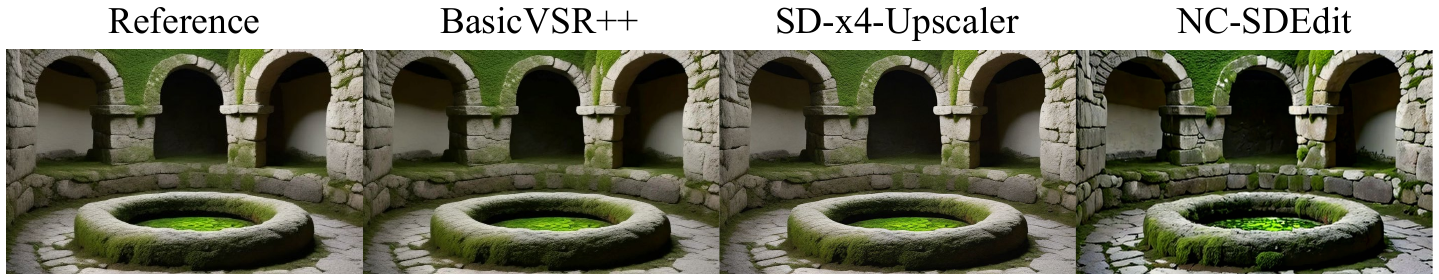}
  \caption{Comparison with entirely different methods
  }
  \label{fig:qita}
\end{figure}

% \begin{figure}[tb]
% \centering
% \begin{minipage}{0.6\linewidth}
% \centering
% \includegraphics[width=\linewidth]{iter.pdf}
% \captionsetup{font=scriptsize}
% \caption{Visual comparisons of different iteration steps $N$}
% \label{fig:fig10}
% \vspace{-11pt}
% \hfill
% \end{minipage}
% \hfill
% \begin{minipage}{0.38\linewidth}
% \centering
% \includegraphics[width=\linewidth]{nu_show.pdf}
% \captionsetup{font=scriptsize}
% \caption{Visual comparisons of different threshold frequency $\nu$}
% \label{fig:fig2}
% %\vspace{-11pt}
% \end{minipage}
% \end{figure}

\begin{figure}[t]
  \centering
  \includegraphics[height=3.8cm]{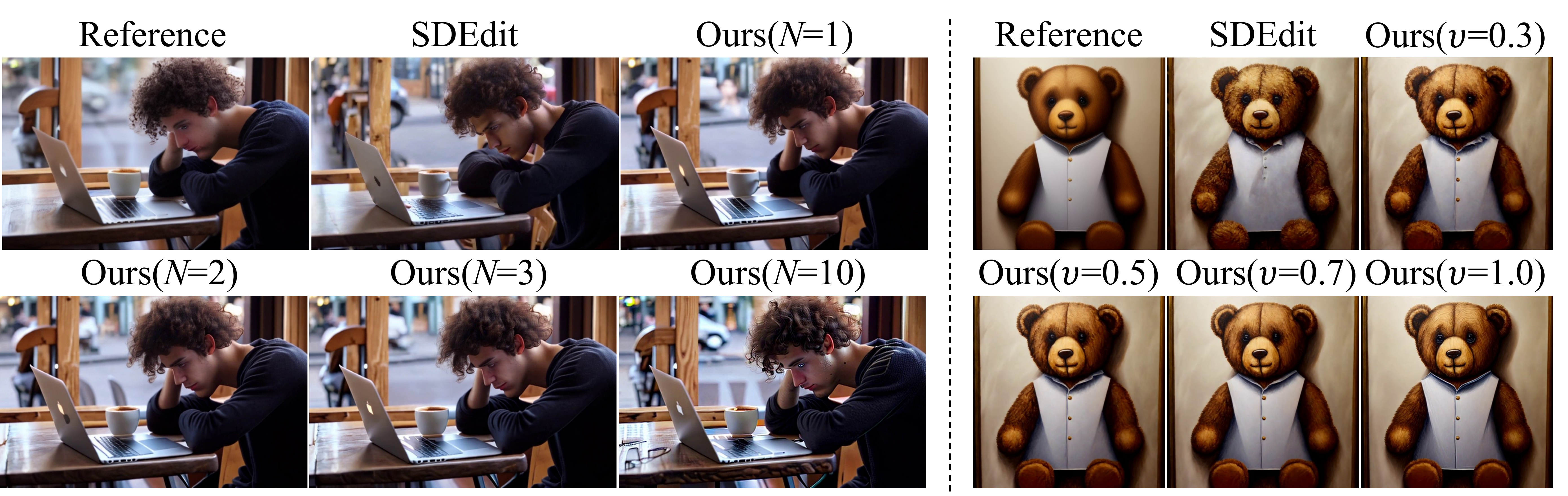}
  \caption{Visual comparisons about iteration steps $N$ and threshold frequency $\nu$
  }
  \label{fig:fig10}
\end{figure}

% \begin{figure}[tb]
%   \centering
%   \includegraphics[height=6cm]{iter.pdf}
%   \caption{Visual comparisons of different iteration steps $N$
%   }
%   \label{fig:fig10}
% \end{figure}

% \begin{figure}[tb]
%   \centering
%   \includegraphics[height=8.2cm]{nu_show.pdf}
%   \caption{Visual comparisons of different threshold frequency $\nu$
%   }
%   \label{fig:nu_show_pdf}
% \end{figure}

\begin{figure}[tb]
  \centering
  \includegraphics[height=2cm]{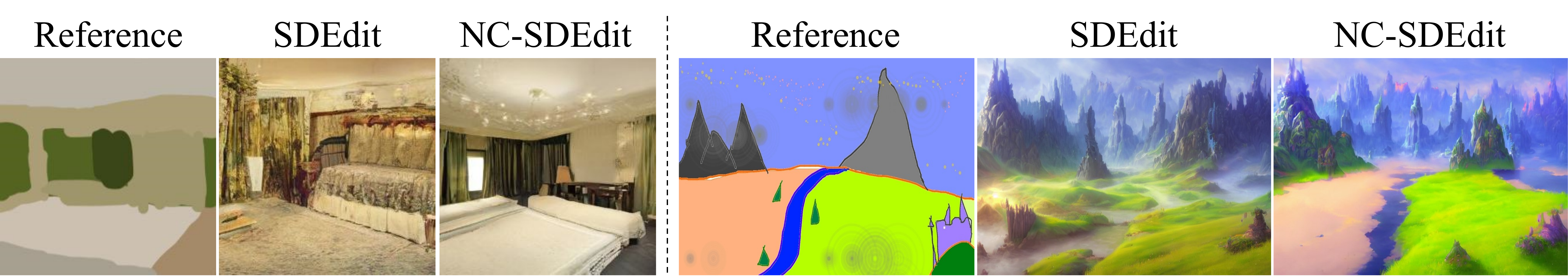}
  \caption{Visual comparisons of Stroke Painting to Image($t_0=800,N=1,\nu=0.1$)}
  \label{fig:strike_pdf}
\end{figure}

\begin{table}[t]
  % \vspace{-1em}
  \caption{Ablation Study about $t_0$ and $\nu$ ($L^l_2$ $\downarrow$/$D_{SF}$ $\uparrow$)
  }
\setlength\tabcolsep{2.6pt}
  \label{tab:table3}
  \centering
  \scalebox{1.0}{
\begin{tabular}{lccccc}
\hline Method & $\nu$=0.0 & $\nu$=0.3 & $\nu$=0.5 & $\nu$=0.7 & $\nu$=1.0  \\
\hline 
$t_0$=500        & 2.9677/0.0544 & 2.2955/0.0535 & 2.0532/0.0548 & $\mathbf{1.9383}$/0.0571  & 2.0706/0.0641 \\
$t_0$=600        & 4.2997/0.0561 & 3.2612/0.0554 & 2.8896/0.0566 & 2.7246/0.0589 & 2.7957/0.0658\\
$t_0$=700        & 6.2563/0.0586 & 4.7217/0.0574 & 4.1862/0.0586  & 3.9240/0.0606 & 3.9010/$\mathbf{0.0664}$\\
\hline
\end{tabular}}
  %\vspace{-2em}
\end{table}

\subsection{Quantitative and Qualitative Evaluation}
\noindent \textbf{Quantitative Evaluation.} We evaluate our method against SDEdit, with the quantitative results presented in \cref{tab:headings}, by enhancing the videos in the reference set at a resolution of 640$\times$1024. According to the results, our proposed method only requires an additional one to three iterations and significantly outperforms previous approaches in all evaluation metrics, with a slight exception in the image quality assessment metric CLIP-IQA. Additionally, it is observed that when the number of iteration steps becomes too large (e.g., 10), the enhancement effect does not increase but rather decreases. We will discuss this in \cref{sec:43}.

\noindent \textbf{Qualitative Evaluation.} \cref{fig:zongshow} presents the visual comparison of video enhancement results with various types and aspect ratios. It can be observed that our method is capable of maintaining the original content during the video enhancement process. In contrast, existing methods either introduce strange noise (as in the 'mosque' case) or object (ike in the 'deer' case), or alter the details of the existing content (e.g., the time in the 'clock' case).

\noindent \textbf{User Study and extra Comparisons.} We also conduct a user study with 18 samples and invite 15 volunteers to evaluate the results. As shown in \cref{tab:userstudy}, our method performs much better than SDEdit, especially in terms of consistency. Additionally, we compare our method with a traditional SR method, BasicVSR++ \cite{2021BasicVSR}, and a SR3-based method, SD-x4-Upscaler \cite{rombach2022high}. Results are shown in \cref{fig:qita}. As discussed in \cref{sec:22}, unlike NC-SDEdit, the two other methods merely increase the resolution and fail to enhance the texture.

\subsection{Ablation Studies and Analysis}
\label{sec:43}
\noindent \textbf{Influence of Iteration Steps $N$.} Our proposed method involves iterating the initial noise multiple times. We provide an intuitive analysis of the effects corresponding to different numbers of iterations using a specific case as an example, depicted in \cref{fig:fig10}.  In the original video, a young man is shown holding his head with one hand while looking at a laptop. However, SDEdit, while enhancing the quality, modifies the content of the original video, such as altering the hand posture and the shape of the coffee cup. In contrast, our method requires only a single iteration to significantly preserve the overall content of the original video while enhancing it. Subsequent iterations, two or three in total, further enhance the preservation of original details, such as the laptop's logo. However, when the number of iteration steps becomes excessive (e.g., 10), the video structure is preserved identically to the reference, but the colors become oversaturated. The reason is that the content loss $||f_{l}^\nu(x^{r})-f_{l}^\nu(\hat{x}_0^{t_0})||$ in \cref{eq:eq9} emphasizes the consistency of low-frequency information, \textit{i.e.,} video structure, which might have side effects on color. 
Designing a more appropriate consistency objective could be a direction to mitigate color oversaturation. As a general guideline, we recommend limiting the iteration steps to between 1 and 3.

\noindent \textbf{Influence of Threshold Frequency $\nu$.} We suggest using the Fourier transform to extract the low-frequency components. The threshold frequency $\nu$ serves as the boundary between high and low frequencies, influencing the constraint range of the objective function $||f_{l}^\nu(x^{r})-f_{l}^\nu(\hat{x}_0^{t_0})||$. Using the 'bear' case in \cref{fig:fig10} as an example, SDEdit, without any additional consistency preservation, can be likened to having a threshold frequency of $\nu=0$ resulting in alterations to details such as the teddy bear's collar button. As the threshold frequency increases, the range constrained by the objective function $||f_{l}^\nu(x^{r})-f_{l}^\nu(\hat{x}_0^{t_0})||$ expands, thereby enhancing the consistency of content between the final enhanced video and the original video. When the threshold frequency reaches 1.0, the texture of the teddy bear's fur in the enhanced video is almost identical to that in the original video. Additionally, the threshold frequency $\nu$ enables the application of our method to other tasks based on SDEdit, such as sketch-to-image translation, as depicted in \cref{fig:strike_pdf}. In addition, as shown in \cref{tab:table3} we conduct an ablation study on 60 videos about initial denoising step $t_0$ and threshold frequency $\nu$. Firstly, it can be observed that increasing $t_0$ results in decreased content consistency but enhances details more effectively. Secondly, for a given $t_0$, a larger $\nu$ improves the effectiveness of content-preserving video enhancement. Ultimately, we choose a compromise parameter combination ($t_0$=600; $\nu$=1.0) as the default hyperparameters.

\begin{figure}[tb]
  \centering
  \includegraphics[height=8cm]{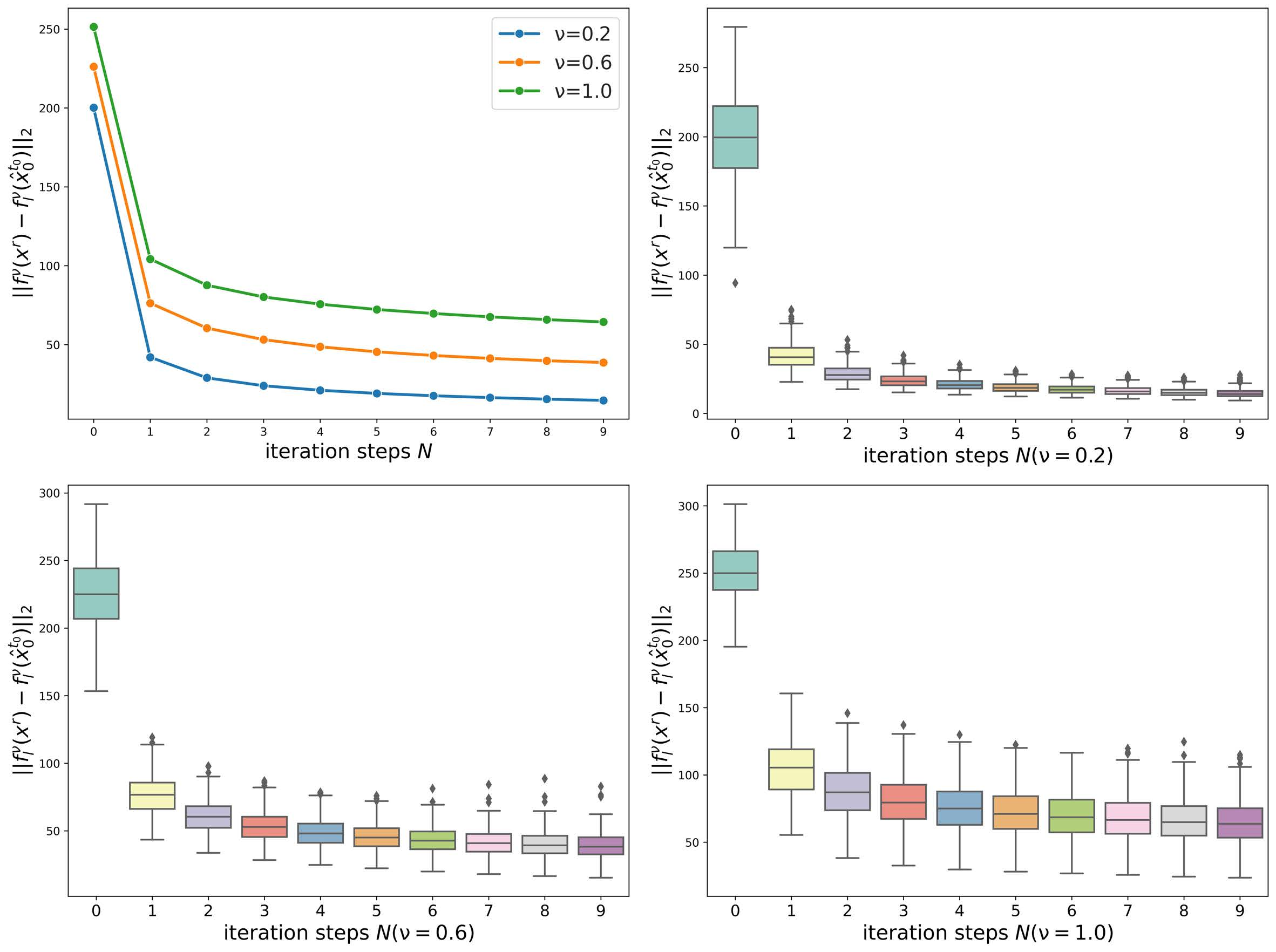}
  \caption{Demonstration of the Optimization Effect of the Objective Function}
  \label{fig:seaborn_plots}
\end{figure}

\begin{figure}[t]
  \centering
  \includegraphics[height=11cm]{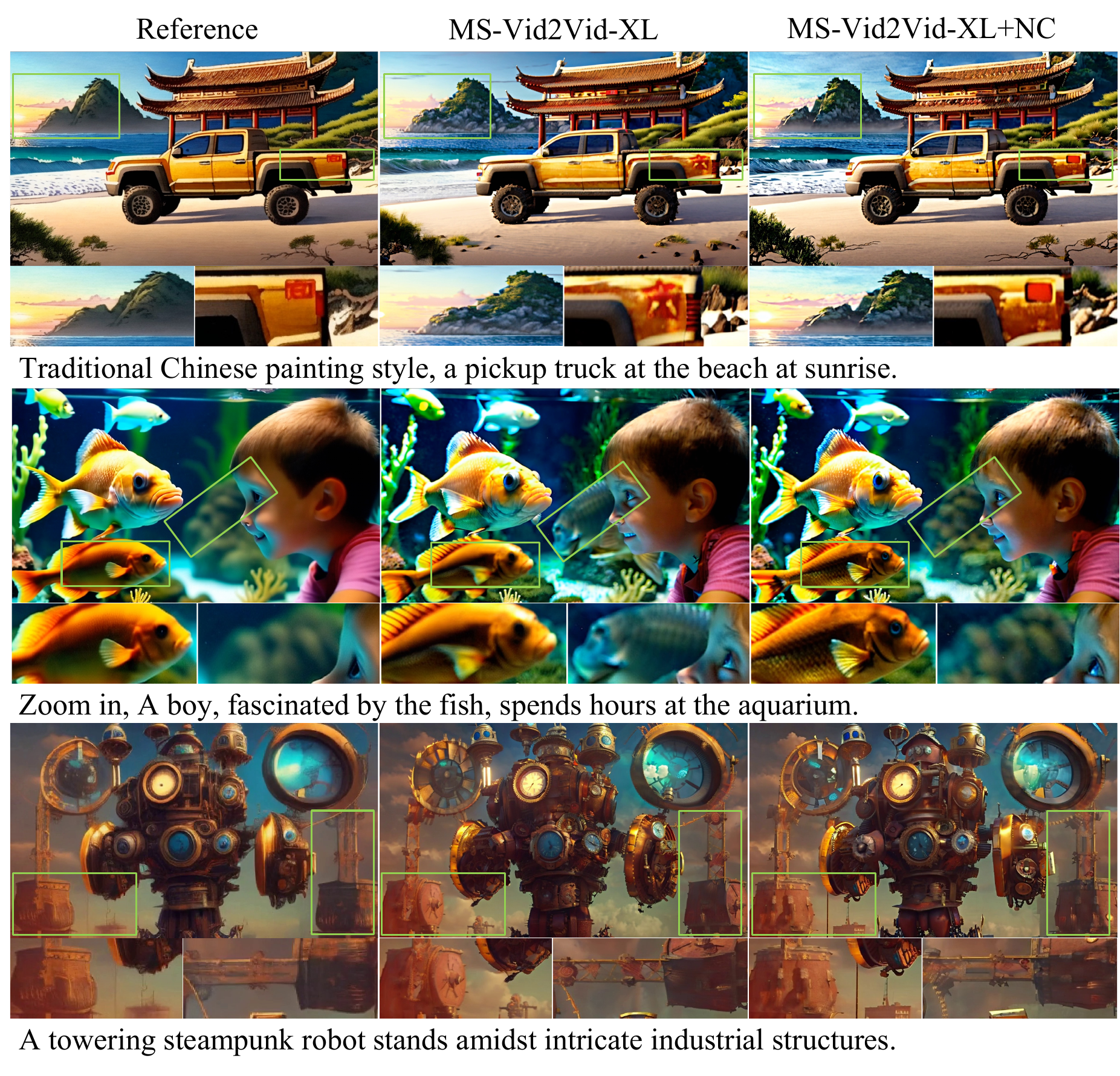}
  \caption{Visual Demonstration of MS-Vid2Vid-XL \cite{zhang2023i2vgen} with Noise Calibration}
  \label{fig:MD_pic_pdf}
\end{figure}

\begin{figure}[t]
  \centering
  \includegraphics[height=8.3cm]{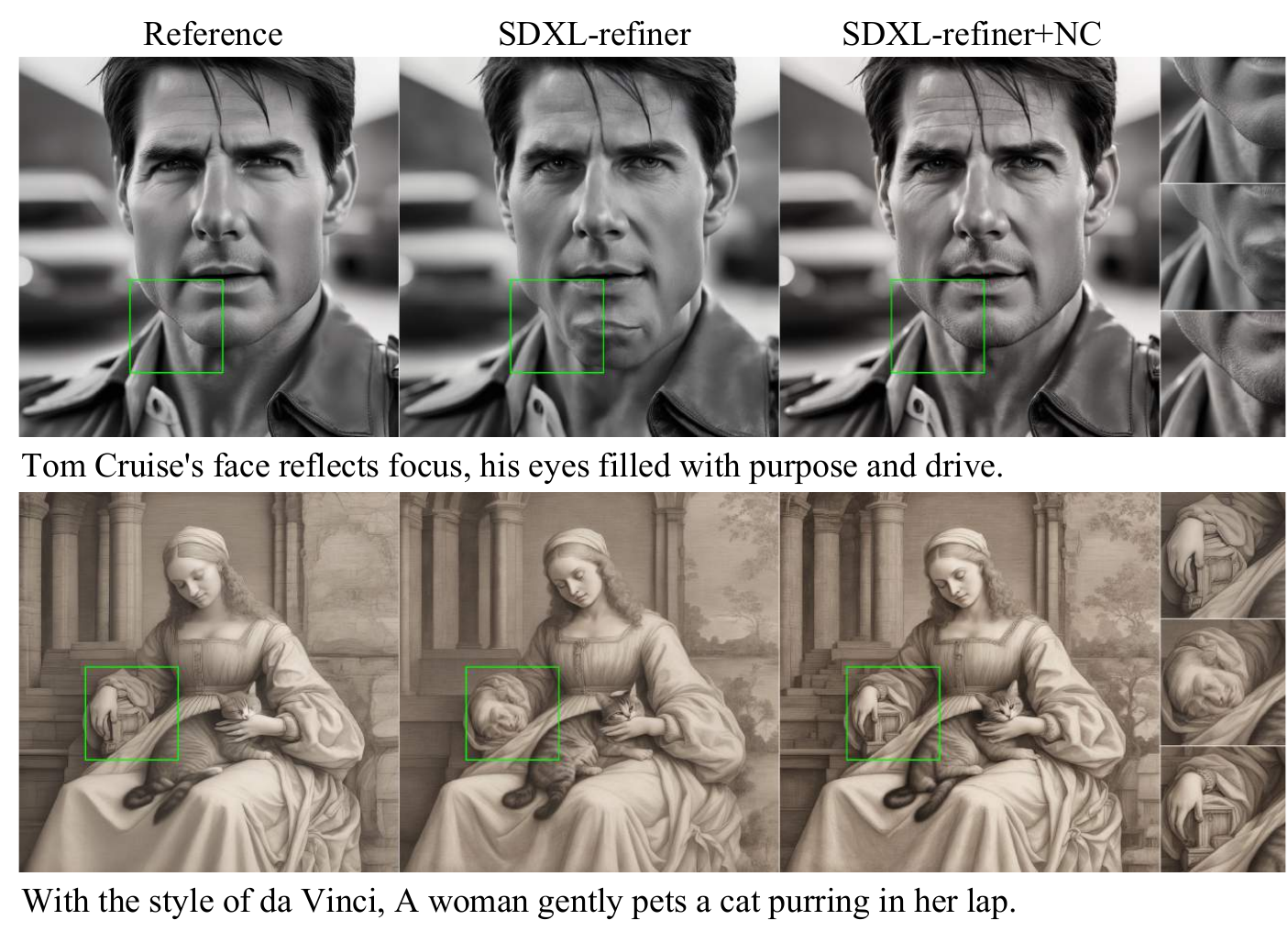}
  \caption{Visual Demonstration of SDXL-1.0-refiner \cite{podell2023sdxl} with Noise Calibration}
  \label{fig:refiner_pdf}
\end{figure}

\begin{table}[t]
  \caption{Quantitative Comparisons based on MS-Vid2Vid-XL and SDXL-1.0-refiner
  }
  \label{tab:tab2}
  \centering
\begin{tabular}{l|ccc|cc|c}
\hline Method& $\mathrm{MSE_l}$$\downarrow$ & MSE$\downarrow$ & SSIM$\uparrow$ & CLIP-IQA$\uparrow$ & DOVER$\uparrow$ & $\mathrm{D_{SF}}$$\uparrow$\\
\hline MS-Vid2Vid-XL \cite{zhang2023i2vgen}& 3.2214 & 0.7490 & 0.7079 & 0.4232 & 52.89 & 0.0478\\
MS-Vid2Vid-XL+NC& $\mathbf{2.6848}$ & $\mathbf{0.7120}$ & $\mathbf{0.7253}$ & $\mathbf{0.4305}$ & $\mathbf{57.61}$ & $\mathbf{0.0517}$ \\
SDXL-1.0-refiner \cite{podell2023sdxl}& 1.2775 & 0.6933 & 0.7344 & 0.8590 & $\text{-}$ & 0.0503\\
SDXL-1.0-refiner+NC& $\mathbf{0.8750}$ & $\mathbf{0.5834}$ & $\mathbf{0.7625}$ & $\mathbf{0.8734 }$ & $\text{-}$ & $\mathbf{0.0530}$\\
\hline
\end{tabular}
\end{table}

\noindent \textbf{Convergence of Noise Calibration.} To further validate whether Noise Calibration can effectively optimize the objective function (\cref{eq:eq9}), We calculate the L2 distance $||f_{l}^\nu(x^{r})-f_{l}^\nu(\hat{x}_0^{t_0})||_2$ in the latent space between $f_{l}^\nu(x^{r})$ and $f_{l}^\nu(\hat{x}_0^{t_0})$ on the Lavie700 dataset \cite{liu2023evalcrafter}, as the number of iterations $N$ increases for different threshold frequencies $\nu$. As shown in \cref{fig:seaborn_plots}, for different values of the threshold frequency $\nu$, the average value of the optimization target decreases as the number of iterations $N$ increases. Notably, the most significant effect is observed with just one iteration, indicating that achieving a high level of consistency maintenance can be accomplished with a single iteration.

\noindent \textbf{Evaluation about training and inference costs.} Firstly, our method is training-free and incurs no training costs. 
Secondly, during inference, using the default setting of initial denoising step $t_0$ = 600 and DDIM steps = 30, our method requires only 1 to 3 additional refinement steps, resulting in less than 10\% additional inference time compared to SDEdit.

\subsection{Improving upon State-of-the-Art Visual Refinement Models}

% \begin{table}[tb]
%   \caption{Quantitative Comparisons based on MS-Vid2Vid-XL and SDXL-1.0-refiner
%   }
%   \label{tab:tab2}
%   \centering
% \begin{tabular}{lcccccccc}
% \hline 
% \multirow{2}*{ Method } & \multicolumn{4}{c}{ MS-Vid2Vid-XL($\nu=1.0$)} & & \multicolumn{3}{c}{ SDXL-1.0-refiner($\nu=0.5$)} \\
% \cline { 2 - 5 } \cline { 7 - 9 } & MSE $\downarrow$ & SSIM $\uparrow$ & CLIP-IQA $\uparrow$ & DOVER $\uparrow$ & & MSE $\downarrow$ & SSIM $\uparrow$ & CLIP-IQA $\uparrow$ \\
% \hline SDEdit & 0.7490 & 0.7079 & 0.4232 & 52.89 & & 0.6933 & 0.7344 & 0.8590 \\
% NC-SDEdit & $\mathbf{0.7120}$ & $\mathbf{0.7253}$ & $\mathbf{0.4305}$ & $\mathbf{57.61}$ & & $\mathbf{0.5834}$ & $\mathbf{0.7625}$ & $\mathbf{0.8734 }$ \\
% \hline
% \end{tabular}
% \end{table}

% Noise Calibration, as a plug-and-play method, can conveniently enhance the performance of existing state-of-the-art visual refinement models.

MS-Vid2Vid-XL \cite{zhang2023i2vgen} and SDXL-1.0-refiner \cite{podell2023sdxl} are two refinement models that fine-tune existing generative models and utilize SDEdit for quality enhancement. However, during the enhancement process, the existing details of the original video/image are often smoothed out, and the originally correct structures tend to be disrupted. Our method, Noise Calibration, can address this issue. To validate the effectiveness of Noise Calibration, we employ the same 700 paired text-video samples from EvalCrafter as the test set for MS-Vid2Vid-XL. Furthermore, we generate corresponding images using SDXL \cite{podell2023sdxl} based on these 700 texts as the test set for SDXL-1.0-refiner. Given the default initial denoising step $t_0=300$ in SDXL-1.0-refiner, we set the threshold frequency $\nu$ to 0.5. The results, as presented in \cref{tab:tab2}, demonstrate that Noise Calibration improves the performance of existing refinement models across all metrics. Moreover, as illustrated in \cref{fig:MD_pic_pdf,fig:refiner_pdf}, our method yields more consistent enhancement effects intuitively.

% \section{Further Discussion}
% \subsubsection{Reproducibility.}
% We will subsequently make our code publicly available.
% \subsubsection{Limitations}
% We are unable to control the degree of consistency of content in different regions of the video. Perhaps this can be improved by using techniques such as image matting.
% \subsubsection{Ethics Statement}
% Our paradigm is based on the pretrained video diffusion model, therefore, it does not introduce any additional ethical concerns.

% \section{Limitation}
% Our method currently lacks the ability to control the degree of consistency of content across different regions of the video. This limitation could potentially be addressed by employing techniques such as image matting.
\section{Limitation, Societal Impact and Acknowledgements}
\noindent \textbf{Limitation.} Like SDEdit, the enhancement effectiveness of our method is also limited by the performance of the base model. 

\noindent \textbf{Societal Impact.} As our method is for improving video quality, it does not introduce additional ethical concerns. 

\noindent \textbf{Acknowledgements.} This research is supported by National Key R\&D Program of China (No. 2018AAA0100300).

\section{Conclusion}
In this work, we propose a novel formulation for video enhancement that takes into account both visual quality and consistency of content. While using the pre-trained T2V diffusion model for denoising to improve video quality, we introduce \textbf{Noise Calibration}, a simple yet effective method for maintaining consistency of content before and after enhancement. Extensive analysis and experiments demonstrate the effectiveness of our approach.

\clearpage  % TODO REVIEW/FINAL: This \clearpage needs to be removed from both review and camera-ready versions.

\bibliographystyle{splncs04}
\bibliography{egbib}
\end{document}